\documentclass[11pt,a4paper]{article}
\usepackage[hyperref]{naaclhlt2019}
\usepackage{times}
\usepackage{latexsym}

\usepackage{amsmath,amssymb,amsfonts}
\usepackage{algorithmic}
\usepackage{graphicx}
\usepackage{textcomp}
\usepackage{xcolor}
\usepackage{multirow}
\usepackage{booktabs}
\usepackage{threeparttable}

\usepackage{url}

\aclfinalcopy 
\def\aclpaperid{951} 

\title{Attentional Encoder Network for Targeted Sentiment Classification}

\author{Youwei Song, Jiahai Wang \thanks{~~The corresponding author.}, Tao Jiang, Zhiyue Liu, Yanghui Rao \\
  School of Data and Computer Science \\
  Sun Yat-sen University \\
  Guangzhou, China \\
  {\tt \{songyw5,jiangt59,liuzhy93\}@mail2.sysu.edu.cn} \\
  {\tt \{wangjiah,raoyangh\}@mail.sysu.edu.cn} \\
}

\date{}

\begin{document}
\maketitle

\begin{abstract}
Targeted sentiment classification aims at determining the sentimental tendency towards specific targets. Most of the previous approaches model context and target words with RNN and attention.
However, RNNs are difficult to parallelize and truncated backpropagation through time brings difficulty in remembering long-term patterns.
To address this issue, this paper proposes an Attentional Encoder Network (AEN) which eschews recurrence and employs attention based encoders for the modeling between context and target.
We raise the label unreliability issue and introduce label smoothing regularization.
We also apply pre-trained BERT to this task and obtain new state-of-the-art results.
Experiments and analysis demonstrate the effectiveness and lightweight of our model.
\footnote{Source code is available at \url{https://github.com/songyouwei/ABSA-PyTorch/tree/aen}.}
\end{abstract}

\section{Introduction}

Targeted sentiment classification is a fine-grained sentiment analysis task, which aims at determining the sentiment polarities (e.g., negative, neutral, or positive) of a sentence over ``opinion targets'' that explicitly appear in the
sentence. For example, given a sentence \textit{``I hated their service, but their food was great''}, the sentiment polarities for the target \textit{``service''} and \textit{``food''} are negative and positive respectively.
A target is usually an entity or an entity aspect.

In recent years, neural network models are designed to automatically learn useful low-dimensional representations from targets and contexts and obtain promising results ~\cite{dong2014adaptive,tang2016effective}.
However, these neural network models are still in infancy to deal with the fine-grained targeted sentiment classification task.

Attention mechanism, which has been successfully used in
machine translation \cite{bahdanau2014neural}, is incorporated to enforce the model to pay more attention to context words with closer semantic relations with the target.
There are already some studies use attention to generate target-specific sentence representations \cite{wang2016attention,ma2017interactive,chen2017recurrent}
or to transform sentence representations according to target words \cite{li2018transformation}.
However, these studies depend on complex recurrent neural networks (RNNs)
as sequence encoder to compute hidden semantics of texts.

The first problem with previous works is that the modeling of text relies on RNNs.
RNNs, such as LSTM, are very expressive, but they are hard to parallelize and backpropagation through time (BPTT) requires large amounts of memory and computation.
Moreover, essentially every training algorithm of RNN is the truncated BPTT, which affects the model's ability to capture dependencies over longer time scales \cite{werbos1990backpropagation}.
Although LSTM can alleviate the vanishing gradient problem to a certain extent and thus maintain long distance information,
this usually requires a large amount of training data.
Another problem that previous studies ignore is the label unreliability issue,
since \textit{neutral} sentiment is a fuzzy sentimental state and brings difficulty for model learning.
As far as we know, we are the first to raise the label unreliability issue in the targeted sentiment classification task.

This paper propose an attention based model to solve the problems above.
Specifically, our model eschews recurrence and employs attention as a competitive alternative to draw the introspective and interactive semantics between target and context words.
To deal with the label unreliability issue, we employ a label smoothing regularization
to encourage the model to be less confident with fuzzy labels.
We also apply pre-trained BERT \cite{devlin2018bert}
to this task and show our model enhances the performance of basic BERT model.
Experimental results on three benchmark datasets show that the proposed model achieves competitive performance and is a lightweight alternative of the best RNN based models.

The main contributions of this work are presented as follows:
\begin{enumerate}
\item We design an attentional encoder network to draw the hidden states and semantic interactions between target and context words.
\item We raise the label unreliability issue and add an effective label smoothing regularization term to the loss function for encouraging the model to be less confident with the training labels.
\item We apply pre-trained BERT to this task, our model enhances the performance of basic BERT model and obtains new state-of-the-art results.
\item We evaluate the model sizes of the compared models and show the lightweight of the proposed model.
\end{enumerate}

\section{Related Work}

The research approach of the targeted sentiment classification task including traditional machine learning methods and neural networks methods.

Traditional machine learning methods, including rule-based methods \cite{ding2008holistic} and statistic-based methods \cite{jiang2011target}, mainly focus on extracting a set of features like sentiment lexicons features and bag-of-words features to train a sentiment classifier \cite{rao2009semi}.
The performance of these methods highly depends on the effectiveness of the feature engineering works, which are labor intensive.

In recent years, neural network methods are getting more and more attention as they do not need handcrafted features and can encode sentences with low-dimensional word vectors where rich semantic information stained.
In order to incorporate target words into a model,
Tang et al. \shortcite{tang2016effective} propose TD-LSTM to extend LSTM by using two single-directional LSTM to model the left context and right context of the target word respectively.
Tang et al. \shortcite{tang2016aspect} design MemNet which consists of a multi-hop attention mechanism with an external memory to capture the importance of each context word concerning the given target. Multiple attention is paid to the memory represented by word embeddings to build higher semantic information.
Wang et al. \shortcite{wang2016attention} propose ATAE-LSTM which concatenates target embeddings with word representations and let targets participate in computing attention weights.
Chen et al. \shortcite{chen2017recurrent} propose RAM which adopts multiple-attention mechanism on the memory built with bidirectional LSTM and nonlinearly combines the attention results with gated recurrent units (GRUs).
Ma et al. \shortcite{ma2017interactive} propose IAN which learns the representations of the target and context with two attention networks interactively.

\section{Proposed Methodology}

Given a context sequence $\mathbf{w^c} = \{w_1^c, w_2^c, ..., w_n^c\}$
and a target sequence $\mathbf{w^t} = \{w_1^t, w_2^t, ..., w_m^t\}$,
where $\mathbf{w^t}$ is a sub-sequence of $\mathbf{w^c}$.
The goal of this model is to predict the sentiment polarity of the
sentence $\mathbf{w^c}$ over the target $\mathbf{w^t}$.

\begin{figure}
\centering
\includegraphics[scale=0.5]{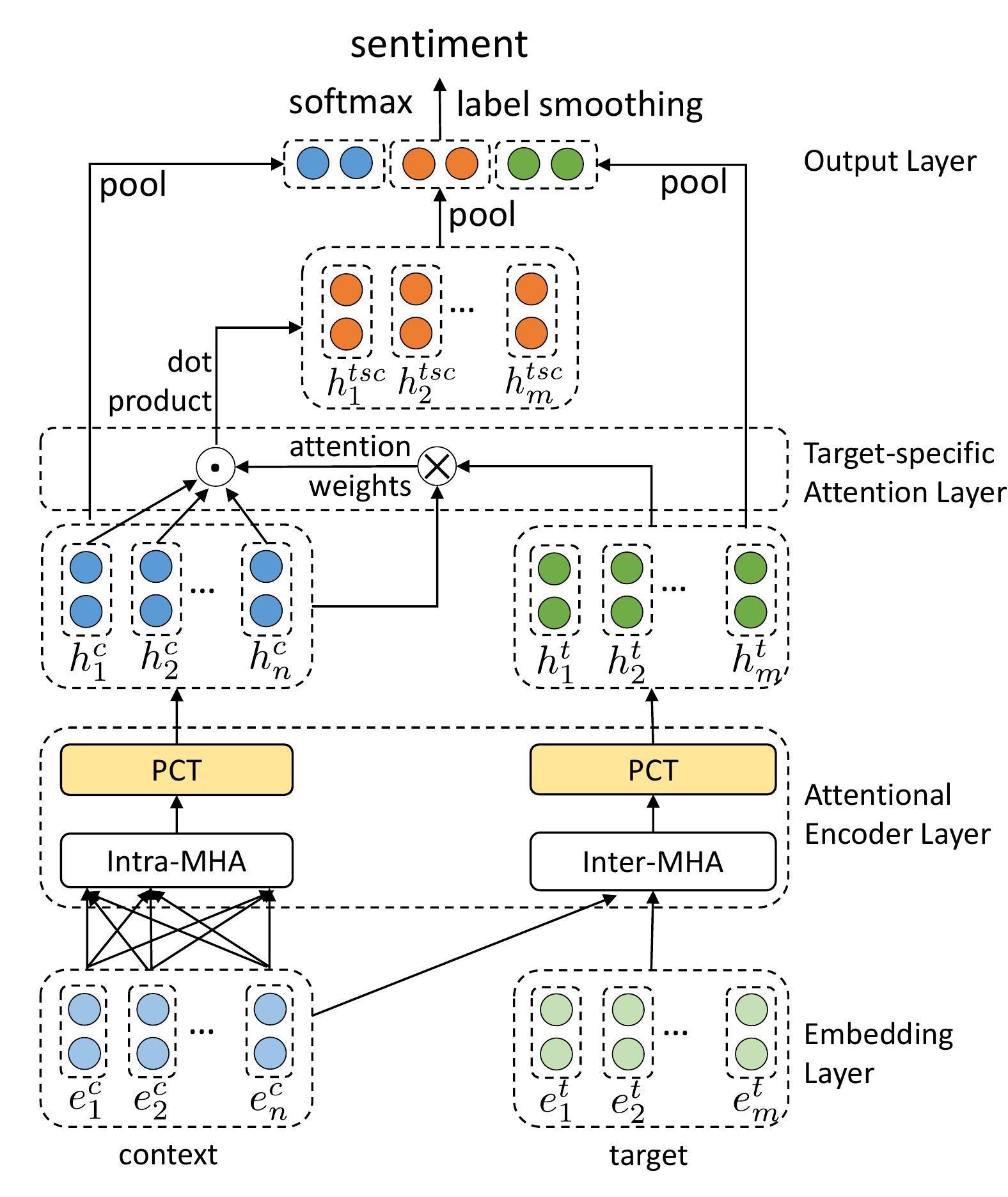}
\caption{Overall architecture of the proposed AEN.}
\label{fig:model}
\end{figure}

Figure \ref{fig:model} illustrates the overall architecture of the proposed \textbf{A}ttentional \textbf{E}ncoder \textbf{N}etwork (AEN), which mainly consists of an embedding layer, an attentional encoder layer, a target-specific attention layer, and an output layer.
Embedding layer has two types: GloVe embedding and BERT embedding.
Accordingly, the models are named \textbf{AEN-GloVe} and \textbf{AEN-BERT}.

\subsection{Embedding Layer}

\subsubsection{GloVe Embedding}

Let $L \in \mathbb{R}^{d_{emb} \times |V|}$ to be the pre-trained GloVe \cite{pennington2014glove} embedding matrix,
where $d_{emb}$ is the dimension of word vectors and $|V|$ is the vocabulary size.
Then we map each word $w^i \in \mathbb{R}^{|V|}$ to its corresponding embedding vector $e_i \in \mathbb{R}^{d_{emb} \times 1}$,
which is a column in the embedding matrix $L$.

\subsubsection{BERT Embedding}

BERT embedding uses the pre-trained BERT to generate word vectors of sequence.
In order to facilitate the training and fine-tuning of BERT model,
we transform the given context and target to
``[CLS] + context + [SEP]'' and ``[CLS] + target + [SEP]'' respectively.

\subsection{Attentional Encoder Layer} \label{Attentional Encoder}

The attentional encoder layer is a parallelizable and interactive alternative of LSTM
and is applied to compute the hidden states of the input embeddings.
This layer consists of two submodules:
the \textbf{M}ulti-\textbf{H}ead \textbf{A}ttention (MHA) and the \textbf{P}oint-wise \textbf{C}onvolution \textbf{T}ransformation (PCT).

\subsubsection{Multi-Head Attention} \label{sec:MHA}

\textbf{M}ulti-\textbf{H}ead \textbf{A}ttention (MHA) is the attention that can perform multiple attention function in parallel.
Different from Transformer \cite{vaswani2017attention}, we use \textbf{Intra-MHA} for introspective context words modeling
and \textbf{Inter-MHA} for context-perceptive target words modeling, which is more lightweight and target is modeled according to a given context.

An attention function maps a key sequence $\mathbf{k} = \{k_1, k_2, ..., k_n\}$ and
a query sequence $\mathbf{q} = \{q_1, q_2, ..., q_m\}$ to an output sequence $\mathbf{o}$:
\begin{align}
Attention(\mathbf{k}, \mathbf{q}) &= softmax(f_{s}(\mathbf{k}, \mathbf{q})) \mathbf{k}
\end{align}
where $f_{s}$ denotes the alignment function which learns the semantic relevance between $q_j$ and $k_i$:
\begin{align}
f_{s}(k_i, q_j) &= tanh([k_i; q_j] \cdot W_{att})
\end{align}
where $W_{att} \in \mathbb{R}^{2d_{hid}}$ are learnable weights.

MHA can learn \emph{n\_head} different scores in parallel child spaces and is very powerful for alignments.
The $n_{head}$ outputs are concatenated and projected to the specified hidden dimension $d_{hid}$, namely,
\begin{align}
MHA(\mathbf{k}, \mathbf{q}) &= [\mathbf{o}^1; \mathbf{o}^2...; \mathbf{o}^{n_{head}}] \cdot W_{mh} \\
\mathbf{o}^h &= Attention^h(\mathbf{k}, \mathbf{q})
\end{align}
where ``$;$'' denotes vector concatenation, $W_{mh} \in \mathbb{R}^{d_{hid} \times d_{hid}}$,
$\mathbf{o}^h = \{o_1^h, o_2^h, ..., o_m^h\}$ is the output of the $h$-th head attention and $h \in [1, n_{head}]$.

\textbf{Intra-MHA}, or multi-head self-attention,
is a special situation for typical attention mechanism that $\mathbf{q} = \mathbf{k}$.
Given a context embedding $\mathbf{e^c}$, we can get the introspective context representation $\mathbf{c^{intra}}$ by:
\begin{align}
\mathbf{c^{intra}} = MHA(\mathbf{e^c}, \mathbf{e^c})
\end{align}
The learned context representation
$\mathbf{c^{intra}}=\{c_1^{intra}, c_2^{intra}, ..., c_n^{intra}\}$ is aware of long-term dependencies.

\textbf{Inter-MHA} is the generally used form of attention mechanism that $\mathbf{q}$ is different from $\mathbf{k}$.
Given a context embedding $\mathbf{e^c}$ and a target embedding $\mathbf{e^t}$,
we can get the context-perceptive target representation $\mathbf{t^{inter}}$ by:
\begin{align}
\mathbf{t^{inter}} = MHA(\mathbf{e^c}, \mathbf{e^t})
\end{align}

After this interactive procedure,
each given target word $e_j^t$ will have a composed representation selected from context embeddings $\mathbf{e^{c}}$.
Then we get the context-perceptive target words modeling $\mathbf{t^{inter}}=\{t_1^{inter}, t_2^{inter}, ..., t_m^{inter}\}$.

\subsubsection{Point-wise Convolution Transformation} \label{sec:PCT}

A \textbf{P}oint-wise \textbf{C}onvolution \textbf{T}
ransformation (PCT)
can transform contextual information gathered by the MHA.
Point-wise means that the kernel sizes are 1 and
the same transformation is applied to every single token belonging to the input.
Formally, given a input sequence $\mathbf{h}$, PCT is defined as:
\begin{align}
PCT(\mathbf{h}) &= \sigma(\mathbf{h} * W_{pc}^1 + b_{pc}^1) * W_{pc}^2 + b_{pc}^2
\end{align}
where $\sigma$ stands for the ELU activation,
$*$ is the convolution operator,
$W_{pc}^1 \in \mathbb{R}^{d_{hid} \times d_{hid}}$ and $W_{pc}^2 \in \mathbb{R}^{d_{hid} \times d_{hid}}$
are the learnable weights of the two convolutional kernels,
$b_{pc}^1 \in \mathbb{R}^{d_{hid}}$ and $b_{pc}^2 \in \mathbb{R}^{d_{hid}}$
are biases of the two convolutional kernels.

Given $\mathbf{c^{intra}}$ and $\mathbf{t^{inter}}$,
PCTs are applied to get the output hidden states of the attentional encoder layer
$\mathbf{h^c}=\{h_1^c, h_2^c, ..., h_n^c\}$
and $\mathbf{h^t}=\{h_1^t, h_2^t, ..., h_m^t\}$
by:
\begin{align}
\mathbf{h^c} &= PCT(\mathbf{c^{intra}}) \\
\mathbf{h^t} &= PCT(\mathbf{t^{inter}})
\end{align}

\subsection{Target-specific Attention Layer}

After we obtain the introspective context representation $\mathbf{h^c}$ and
the context-perceptive target representation $\mathbf{h^t}$,
we employ another MHA to obtain the target-specific context representation $\mathbf{h^{tsc}}=\{h_1^{tsc}, h_2^{tsc}, ..., h_m^{tsc}\}$ by:
\begin{align}
\mathbf{h^{tsc}} = MHA(\mathbf{h^c}, \mathbf{h^t})
\end{align}
The multi-head attention function here also has its independent parameters.

\subsection{Output Layer}

We get the final representations of the previous outputs by average pooling,
concatenate them as the final comprehensive representation $\mathbf{\tilde{o}}$,
and use a full connected layer to project the concatenated vector into the space of the targeted $C$ classes.
\begin{align}
\mathbf{\tilde{o}} &= [h_{avg}^c; h_{avg}^t; h_{avg}^{tsc}] \\
x &= \tilde{W_o}^T{{\mathbf{\tilde{o}}}}+\tilde{b_o} \\
y &= softmax(x) \\
    &= \frac{exp(x)}{\sum_{k=1}^{C} exp(x)}
\end{align}
where $y \in \mathbb{R}^{C}$ is the predicted sentiment polarity distribution,
$\tilde{W_o} \in \mathbb{R}^{1 \times C}$ and $\tilde{b_o} \in \mathbb{R}^{C}$ are learnable parameters.

\subsection{Regularization and Model Training} \label{sec:LSR}

Since \textit{neutral} sentiment is a very fuzzy sentimental state, training samples which labeled \textit{neutral} are unreliable.
We employ a \textbf{L}abel \textbf{S}moothing \textbf{R}egularization (LSR) term in the loss function.
which penalizes low entropy output distributions \cite{szegedy2016rethinking}.
LSR can reduce overfitting by preventing a network from assigning the full probability to each training example during training, replaces the 0 and 1 targets for a classifier with smoothed values like 0.1 or 0.9.

For a training sample $x$ with the original ground-truth label distribution $q(k|x)$,
we replace $q(k|x)$ with
\begin{align}
q(k|x) = (1-\epsilon) q(k|x) + \epsilon u(k)
\end{align}
where $u(k)$ is the prior distribution over labels ,
and $\epsilon$ is the smoothing parameter.
In this paper, we set the prior label distribution to be uniform $u(k) = 1/C$.

LSR is equivalent to the KL divergence between the prior label distribution $u(k)$ and the network's predicted distribution $p_\theta$.
Formally, LSR term is defined as:
\begin{align}
\mathcal{L}_{lsr} = - D_{KL}(u(k) \| p_\theta)
\end{align}

The objective function (loss function) to be optimized is the cross-entropy loss with $\mathcal{L}_{lsr}$ and $\mathcal{L}_2$ regularization, which is defined as:

\begin{align}
\mathcal{L}(\theta) = - \sum_{i=1}^{C} \hat{y}^c log (y^c) + \mathcal{L}_{lsr} + \lambda \sum_{\theta \in \Theta} {\theta}^2 &
\end{align}
where $\hat{y} \in \mathbb{R}^C $ is the ground truth represented as a one-hot vector,
$y$ is the predicted sentiment distribution vector given by the output layer,
$\lambda$ is the coefficient for $\mathcal{L}_2$ regularization term, and $\Theta$ is the parameter set.


\section{Experiments}

\subsection{Datasets and Experimental Settings}

We conduct experiments on three datasets: SemEval 2014 Task 4 \footnote{The detailed introduction of this task can be found at \url{http://alt.qcri.org/semeval2014/task4}.} \cite{pontiki2014semeval} dataset composed of \emph{Restaurant} reviews and \emph{Laptop} reviews, and ACL 14 \emph{Twitter} dataset gathered by Dong et al. \shortcite{dong2014adaptive}. These datasets are labeled with three sentiment polarities: \emph{positive}, \emph{neutral} and \emph{negative}.
Table \ref{tab:stat} shows the number of training and test instances in each category.

Word embeddings in AEN-GloVe do not get updated in the learning process,
but we fine-tune pre-trained BERT
\footnote{We use uncased BERT-base from \url{https://github.com/google-research/bert}.} in AEN-BERT.
Embedding dimension $d_{dim}$ is 300 for GloVe and is 768 for pre-trained BERT.
Dimension of hidden states $d_{hid}$ is set to 300.
The weights of our model are initialized with Glorot initialization \cite{glorot2010understanding}.
During training, we set label smoothing parameter $\epsilon$ to 0.2 \cite{szegedy2016rethinking}, the coefficient $\lambda$ of $\mathcal{L}_2$ regularization item is $10^{-5}$ and dropout rate is 0.1.
Adam optimizer \cite{kingma2014adam} is applied to update all the parameters.
We adopt the \emph{Accuracy} and \emph{Macro-F1} metrics to evaluate the performance of the model.

\begin{table}[tp]
  \small
  \centering
  \begin{threeparttable}
  \caption{Statistics of the datasets.}
    \begin{tabular}{ccccccc}
    \toprule
    \multirow{2}{*}{\textbf{Dataset}}&
    \multicolumn{2}{c}{\textbf{Positive}}&\multicolumn{2}{c}{\textbf{Neural}}&\multicolumn{2}{c}{\textbf{Negative}}\cr
    \cmidrule(lr){2-3} \cmidrule(lr){4-5} \cmidrule(lr){6-7}
    &Train&Test&Train&Test&Train&Test \cr
    \midrule
        Twitter     &1561 &173 &3127 &346 &1560 &173 \cr
        Restaurant  &2164 &728 &637 &196 &807 &196 \cr
        Laptop      &994 &341 &464 &169 &870 &128 \cr
    \bottomrule
    \end{tabular}
    \label{tab:stat}
    \end{threeparttable}
\end{table}

\subsection{Model Comparisons}

In order to comprehensively evaluate and analysis the performance of AEN-GloVe,
we list 7 baseline models and design 4 ablations of AEN-GloVe.
We also design a basic BERT-based model to evaluate the performance of AEN-BERT.

~\\
\textbf{Non-RNN based baselines:}

$\bullet$ \textbf{Feature-based SVM} \cite{kiritchenko2014nrc} is a traditional support vector machine based model with extensive feature engineering.

$\bullet$ \textbf{Rec-NN} \cite{dong2014adaptive} firstly uses rules to transform the dependency tree and put the opinion target at the root, and then
learns the sentence representation toward target via semantic composition using Recursive NNs.

$\bullet$ \textbf{MemNet} \cite{tang2016aspect} uses multi-hops of attention layers on the context word embeddings for sentence representation to explicitly captures the importance of each context word.

~\\
\textbf{RNN based baselines:}

$\bullet$ \textbf{TD-LSTM} \cite{tang2016effective} extends LSTM by using two LSTM networks to model the left context with target and the right context with target respectively. The left and right target-dependent representations are concatenated for predicting the sentiment polarity of the target.

$\bullet$ \textbf{ATAE-LSTM} \cite{wang2016attention} strengthens the effect of target embeddings, which appends the target embeddings with each word embeddings and use LSTM with attention to get the final representation for classification.

$\bullet$ \textbf{IAN} \cite{ma2017interactive} learns the representations of the target and context with two LSTMs and attentions interactively, which generates the representations for targets and contexts with respect to each other.

$\bullet$ \textbf{RAM} \cite{chen2017recurrent} strengthens MemNet by representing memory with bidirectional LSTM and using a gated recurrent unit network to combine the multiple attention outputs for sentence representation.

~\\
\textbf{AEN-GloVe ablations:}

$\bullet$ \textbf{AEN-GloVe w/o PCT} ablates PCT module.

$\bullet$ \textbf{AEN-GloVe w/o MHA} ablates MHA module.

$\bullet$ \textbf{AEN-GloVe w/o LSR} ablates label smoothing regularization.

$\bullet$ \textbf{AEN-GloVe-BiLSTM} replaces the attentional encoder layer with two bidirectional LSTM.

~\\
\textbf{Basic BERT-based model:}

$\bullet$ \textbf{BERT-SPC} feeds sequence ``[CLS] + context + [SEP] + target + [SEP]''
into the basic BERT model for sentence pair classification task.

\subsection{Main Results}

\begin{table*}[tp]
  \small
  \centering
  \begin{threeparttable}
  \caption{Main results.
  The results of baseline models are retrieved from published papers.
  ``-" means not reported.
  Top 3 scores are in \textbf{bold}.}
    \begin{tabular}{cccccccc}
    \toprule
    \multirow{2}{*}{ }&\multirow{2}{*}{\textbf{Models}}&
    \multicolumn{2}{c}{\textbf{Twitter}}&\multicolumn{2}{c}{\textbf{Restaurant}}&\multicolumn{2}{c}{\textbf{Laptop}}\cr
    \cmidrule(lr){3-4} \cmidrule(lr){5-6} \cmidrule(lr){7-8}
    &&Accuracy&Macro-F1&Accuracy&Macro-F1&Accuracy&Macro-F1\cr
    \midrule
        \multirow{4}*{\textbf{RNN baselines}}
        &TD-LSTM           &0.7080&0.6900              &0.7563&-                  &0.6813&-         \cr
        &ATAE-LSTM         &-&-                        &0.7720&-                  &0.6870&-         \cr
        &IAN               &-&-                        &0.7860&-                  &0.7210&-         \cr
        &RAM               &0.6936&0.6730              &0.8023&0.7080             &\textbf{0.7449}&\textbf{0.7135}     \cr
    \midrule
        \multirow{3}*{\textbf{Non-RNN baselines}}
        &Feature-based SVM &0.6340&0.6330              &0.8016&-                  &0.7049&-           \cr
        &Rec-NN            &0.6630&0.6590              &-&-                       &-&-              \cr
        &MemNet            &0.6850&0.6691              &0.7816&0.6583             &0.7033&0.6409    \cr
    \midrule
        \multirow{4}*{\textbf{AEN-GloVe ablations}}
        &AEN-GloVe w/o PCT       &0.7066&0.6907              &0.8017&0.7050             &0.7272&0.6750 \cr
        &AEN-GloVe w/o MHA       &0.7124&0.6953              &0.7919&0.7028             &0.7178&0.6650 \cr
        &AEN-GloVe w/o LSR       &0.7080&0.6920              &0.8000&0.7108             &0.7288&0.6869 \cr
        &AEN-GloVe-BiLSTM        &0.7210&\textbf{0.7042}     &0.7973&0.7037             &0.7312&0.6980 \cr
    \midrule
        \multirow{3}*{\textbf{Ours}}
        &AEN-GloVe  &\textbf{0.7283}&0.6981  &\textbf{0.8098}&\textbf{0.7214}  &0.7351&0.6904 \cr
        &BERT-SPC  &\textbf{0.7355}&\textbf{0.7214} &\textbf{0.8446}&\textbf{0.7698} &\textbf{0.7899}&\textbf{0.7503} \cr
        &AEN-BERT &\textbf{0.7471}&\textbf{0.7313} &\textbf{0.8312}&\textbf{0.7376} &\textbf{0.7993}&\textbf{0.7631} \cr
    \bottomrule
    \end{tabular}
    \label{tab:result}
    \end{threeparttable}
\end{table*}

Table \ref{tab:result} shows the performance comparison of AEN with other models.
BERT-SPC and AEN-BERT obtain substantial accuracy improvements,
which shows the power of pre-trained BERT on small-data task.
The overall performance of AEN-BERT is better than BERT-SPC,
which suggests that it is important to design a downstream network customized to a specific task.
As the prior knowledge in the pre-trained BERT is not specific to any particular domain,
further fine-tuning on the specific task is necessary for releasing the true power of BERT.

The overall performance of TD-LSTM is not good since it only makes a rough treatment of the target words.
ATAE-LSTM, IAN and RAM are attention based models, they stably exceed the TD-LSTM method on \emph{Restaurant} and \emph{Laptop} datasets.
RAM is better than other RNN based models, but it does not perform well on \emph{Twitter} dataset,
which might because bidirectional LSTM is not good at modeling small and ungrammatical text.

Feature-based SVM
is still a competitive baseline,
but relying on manually-designed features.
Rec-NN gets the worst performances among all neural network baselines
as dependency parsing is not guaranteed to work well on ungrammatical short texts such as tweets and comments.
Like AEN, MemNet also eschews recurrence, but its overall performance is not good
since it does not model the hidden semantic of embeddings, and the result of the last attention is essentially a linear combination of word embeddings.


\subsection{Model Analysis}

As shown in Table \ref{tab:result}, the performances of AEN-GloVe ablations are incomparable
with AEN-GloVe in both accuracy and macro-F1 measure.
This result shows that all of these discarded components are crucial for a good performance.
Comparing the results of AEN-GloVe and AEN-GloVe w/o LSR, we observe that the accuracy of AEN-GloVe w/o LSR drops significantly on all three datasets.
We could attribute this phenomenon to the unreliability of the training samples with \textit{neutral} sentiment.
The overall performance of AEN-GloVe and AEN-GloVe-BiLSTM is relatively close,
AEN-GloVe performs better on the \emph{Restaurant} dataset.
More importantly, AEN-GloVe has fewer parameters and is easier to parallelize.

To figure out whether the proposed AEN-GloVe is a lightweight alternative of recurrent models, we study the model size of each model on the \emph{Restaurant} dataset.
Statistical results are reported in Table \ref{tab:result2}.
We implement all the compared models base on the same source code infrastructure,
use the same hyperparameters, and run them on the same GPU
\footnote{NVIDIA GTX 1080ti. }.

RNN-based and BERT-based models indeed have larger model size.
ATAE-LSTM, IAN, RAM, and AEN-GloVe-BiLSTM are all attention based RNN models,
memory optimization for these models will be more difficult
as the encoded hidden states must be kept simultaneously in memory in order to perform attention mechanisms.
MemNet has the lowest model size as it only has one shared attention layer and two linear layers, it does not calculate hidden states of word embeddings.
AEN-GloVe's lightweight level ranks second,
since it takes some more parameters than MemNet in modeling hidden states of sequences.
As a comparison, the model size of AEN-GloVe-BiLSTM is more than twice that of AEN-GloVe, but does not bring any performance improvements.

\begin{table}[tp]
    \small
    \centering
    \caption{Model sizes. Memory footprints are evaluated on the Restaurant dataset. Lowest 2 are in \textbf{bold}.}
    \begin{tabular}{ccc}
    \toprule
    \multirow{2}{*}{\textbf{Models}}&
    \multicolumn{2}{c}{\textbf{Model size}}\cr
    \cmidrule(lr){2-3}
    &Params $\times 10^6$ & Memory (MB) \cr
    \midrule
    TD-LSTM      &1.44 &12.41\\
    ATAE-LSTM    &2.53 &16.61\\
    IAN          &2.16 &15.30\\
    RAM          &6.13 &31.18\\
    MemNet       &\textbf{0.36} &\textbf{7.82}\\
    \midrule
    AEN-BERT     &112.93 &451.84\\
    AEN-GloVe-BiLSTM   &3.97 &22.52\\
    AEN-GloVe          &\textbf{1.16} &\textbf{11.04}\\
    \bottomrule
    \end{tabular}
    \label{tab:result2}
\end{table}

\section{Conclusion}

In this work, we propose an attentional encoder network for the targeted sentiment classification task.
which employs attention based encoders for the modeling between context and target.
We raise the the label unreliability issue add a label smoothing regularization
to encourage the model to be less confident with fuzzy labels.
We also apply pre-trained BERT to this task and obtain new state-of-the-art results.
Experiments and analysis demonstrate the effectiveness and lightweight of the proposed model.

\bibliographystyle{acl_natbib}
\bibliography{refs}

\end{document}